\documentclass[sigconf]{acmart}

\AtBeginDocument{%
  }

\setcopyright{acmlicensed}
\copyrightyear{2018}
\acmYear{2018}
\acmDOI{XXXXXXX.XXXXXXX}
\acmConference[Conference acronym 'XX]{Make sure to enter the correct
  conference title from your rights confirmation email}{June 03--05,
  2018}{Woodstock, NY}
\acmISBN{978-1-4503-XXXX-X/2018/06}

\usepackage[utf8]{inputenc} 
\usepackage[T1]{fontenc}    
\usepackage{hyperref}       
\usepackage{url}            
\usepackage{booktabs}       
\usepackage{amsfonts}       
\usepackage{nicefrac}       
\usepackage{microtype}      
\usepackage{xcolor}         
\usepackage{makecell}
\usepackage{multirow}

\usepackage{amsmath}
\usepackage{mathtools}
\usepackage{amsthm}
\usepackage{diagbox}
\usepackage{tabularx}

\usepackage{bm}
\usepackage{graphicx}
\usepackage{multirow}
\usepackage{colortbl}
\usepackage{rotating}
\usepackage{makecell}
\usepackage{wrapfig}
\usepackage{enumitem}
\usepackage{pifont}
\usepackage{makecell}
\usepackage[ruled,vlined,linesnumbered]{algorithm2e}

\usepackage[most]{tcolorbox}

\usepackage{multicol}
\usepackage{setspace}

\definecolor{best}{rgb}{1,  0,  0}
\definecolor{second}{rgb}{0,  0,  0.859}
\usepackage{hyperref}
\SetCommentSty{footnotesize}
\SetKwComment{Comment}{$\triangleright$ }{}
\SetSideCommentRight

\begin{document}

\title{Think Before You Act: Intention-Guided Reasoning for LLM-Based Location Prediction}

\author{Qingxiang Liu}
\affiliation{%
  \institution{The Hong Kong University of Science and Technology (Guangzhou)}
  \country{China}
}

\author{Anqi Liang}
\affiliation{
\institution{Shanghai Jiao Tong University}
  \country{China}
}

\author{Zhuoyang Jiang}
\affiliation{%
  \institution{The Hong Kong University of Science and Technology (Guangzhou)}
  \country{China}
}

\author{Yutian Jiang}
\affiliation{%
  \institution{The Hong Kong University of Science and Technology (Guangzhou)}
  \country{China}
}

\author{Sisuo Lyu}
\affiliation{
\institution{The Hong Kong University of Science and Technology}
  \country{Hong Kong}
}

\author{Yu Ji}
\affiliation{
\institution{Fudan University}
\country{China}
}

\author{Haomin Wen}
\affiliation{
\institution{Shanghai Innovation Institute}
\country{China}
}

\author{Yuxuan Liang}
\authornote{Corresponding Author. Email: yuxliang@outlook.com}
\affiliation{%
  \institution{The Hong Kong University of Science and Technology (Guangzhou)}
  \country{China}
}

\renewcommand{\shortauthors}{Liu et al.}

\begin{abstract}
Predicting a user's next Point-of-Interest (POI) based on their historical check-in records is a fundamental task in location-based services. 
While recent methods incorporating large language models have shown strong reasoning capabilities and promising results, they typically formulate the prediction task as a one-step trajectory-to-location mapping problem, making predictions prone to shallow trajectory correlations and historical frequency bias.
We argue that users rarely choose locations directly and instead, they usually first form a traveling intention and then accordingly select specific POIs. Motivated by this insight, we propose \textbf{IntentPOI}, a two-stage intention-guided reasoning framework. 
In the \textit{thinking} stage, we infer users' intermediate intentions by incorporating historical mobility patterns, similar peer behaviors, and the temporal contexts.
In the \textit{acting} stage, we first construct a compact candidate pool, and then perform intention-guided reasoning to identify locations that best align with the inferred intention. 
By explicitly decoupling intention inference from location prediction, {IntentPOI} transforms the next POI prediction from direct trajectory matching into intention-guided reasoning.
Extensive experiments on three real-world datasets demonstrate that {IntentPOI} consistently outperforms eleven state-of-the-art baselines.

\end{abstract}



\received{20 February 2007}
\received[revised]{12 March 2009}
\received[accepted]{5 June 2009}

\maketitle

\section{Introduction}


The rapid development of urban computing and location-based services has promoted Point-of-Interest- (POI) aware applications in diverse real-world scenarios, including route planning, targeted advertising, and trajectory prediction~\cite{luca2021survey,chen2025self}.
As a fundamental task, next POI prediction aims to recommend the location a user is likely to visit at a given time point by analyzing mobility patterns from historical check-in trajectories~\cite{zhao2020go,lai2024disentangled}.
The deep learning-based next POI prediction methods predominantly model sequential check-in records through recurrent networks~\cite{zhao2020go,feng2018deepmove,wu2020personalized}, attention mechanisms~\cite{luo2021stan,sun2024going,yang2022getnext}, or graph neural networks~\cite{lim2020stp,10.1145/3510409} to capture spatial and temporal transition patterns~\cite{dang2023uniform,yin2023next,zeng2025global,wu2025beyond}.
While achieving promising performance, these methods heavily rely on implicit pattern matching over historical trajectories and lack the underlying analysis, which limits their generalization and interpretability in complex urban environments~\cite{wang2024embracing,yang2024siamese}.

Inspired by the prominent reasoning capabilities of Large Language Models (LLMs), recent researchers have applied LLMs to next POI prediction~\cite{wu2025mas4poi,Comapoi,LLM4POI,LLMmove,QT-Mob,lv2026reasoning,tan2024idgenrec}, yielding two main paradigms.
Prompt-based methods reorganize historical trajectories into textual prompts, guiding LLMs to infer user profiles and mobility patterns for evidence-based prediction~\cite{Comapoi,LLM4POI,LLMmove}.
For example, CoMaPOI prompts LLMs with historical trajectories to derive long-term user profiles, short-term mobility patterns, and candidate POIs for final re-ranking.
Token-based methods pretrain semantic tokens for POIs, allowing LLMs to directly learn POI-level knowledge in textual space~\cite{QT-Mob,lv2026reasoning}.
For instance, QT-Mob encodes each POI's overview, geographic location, and spatial context into four discrete tokens and trains the LLM to map these combined tokens to the exact POI index.
These two types of methods demonstrate that LLMs can effectively capture mobility patterns and achieve notable improvements over deep learning-based approaches.

Despite their promising performance, existing LLM-based methods tend to treat the next POI prediction as a \textit{trajectory-to-location mapping problem}, where the adopted LLMs directly predict the targeted POI or re-rank candidates based on the given historical check-in records, thus prone to shallow trajectory correlations and historical frequency bias.
This limitation becomes particularly severe when historical trajectories are sparse or multiple candidates exhibit similar transition patterns. In such cases, the LLM often defaults to frequently visited locations, even though these predictions are semantically inconsistent with the current context. 
In fact, \textit{users rarely make mobility decisions directly at the POI level}. Instead, they typically form \textbf{explicit traveling intentions} (such as dining, shopping, or socializing) and then choose specific locations that satisfy their intentions under spatial and temporal constraints.
This insight suggests that the next POI prediction should not be formulated as a single-step forward problem but as a two-stage reasoning process that \textit{first infers the user's intention and then selects intention-aligned locations}.

\begin{figure}[!t]
	\centering
    \includegraphics[width=\columnwidth]{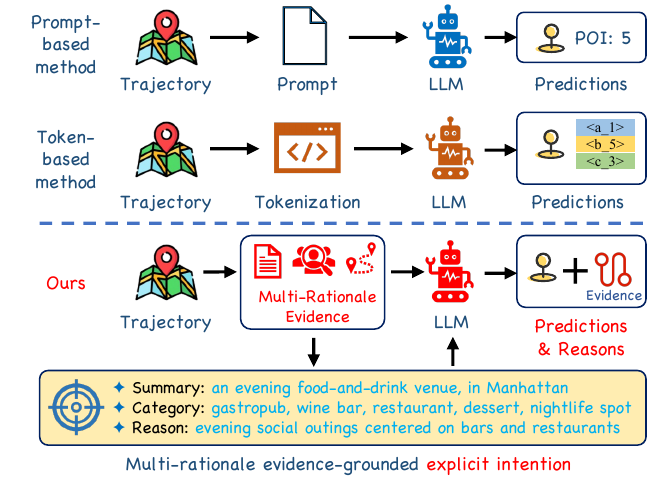}
	\caption{Both the prompt- and token-based methods perform one-step prediction for the next POI index or tokens.
    Our proposed \textbf{IntentPOI} explicitly infers the user's intention as an intermediate step, which serves as a reasoning scaffold to guide the downstream LLM toward intention-aligned reasoning.
    }
    \label{fig_intro_comp}
\end{figure}


To address this gap, we propose a \textbf{thinking-before-acting} principle for LLM-based next POI prediction. Instead of directly mapping historical trajectories to locations, the next POI prediction problem is decomposed into two reasoning stages. The \textit{thinking} stage focuses on understanding the reason \emph{why} the user is likely to travel by explicitly inferring latent intention from historical mobility patterns and contextual signals. The \textit{acting} stage focuses on determining \emph{where} the user will go by recommending locations that best satisfy the inferred intention. The explicit inferred intention serves as a reasoning scaffold that bridges mobility understanding and downstream POI prediction.


Building upon this principle, we propose \textbf{IntentPOI}, a two-stage intention-guided LLM reasoning framework for next POI prediction. In the \textit{thinking} stage, IntentPOI integrates multi-rationale evidence, including user profiles, peer behaviors, and temporal contexts, to infer the user's latent intention through LLM reasoning. 
In the \textit{acting} stage, IntentPOI first constructs a compact candidate pool by combining historically visited POIs with spatially proximate POIs, and then performs intention-guided reasoning to identify locations that best align with the inferred intention. Through this two-stage process, IntentPOI transforms next POI prediction from frequency-driven trajectory matching into intention-grounded reasoning.

Our contributions are summarized as follows:
\begin{itemize}[leftmargin=*]
\item We identify the lack of explicit intention modeling as a fundamental limitation of existing LLM-based next POI prediction methods and reformulate next POI prediction as a two-stage reasoning problem consisting of intention inference and location determination.
\item We propose {IntentPOI}, a thinking-before-acting LLM reasoning framework, to infer the intentions based on multi-rationale evidence and then perform intention-guided candidate recommendation.
\item Extensive experiments on three real-world datasets demonstrate the effectiveness and efficiency of IntentPOI compared with state-of-the-art baselines.
\end{itemize}

\section{Related Work}

\textbf{Deep learning-based approaches.} 
Early deep learning-based next POI prediction methods commonly employed sequential architectures, including RNN-based and attention-based models, to model users' check-in sequences, thereby capturing their mobility patterns and inferring location preferences \cite{Deepmove, Vanext,Flashback,Mobtcast,TSPN-RA}. For example, DeepMove \cite{Deepmove} adopts an RNN-based architecture that incorporates multiple mobility-related factors to model human transition regularities, and introduces a historical attention mechanism to capture periodic patterns from users’ long-term mobility histories. 
Flashback \cite{Flashback} mitigates trajectory sparsity by enabling RNNs to selectively revisit relevant past states. It uses spatial and temporal signals to weight previous trajectory representations, thereby improving next POI prediction performance. MobTCast \cite{Mobtcast} is a context-aware Transformer-based model that incorporates spatio-temporal, semantic, social, and geographic contexts for next POI prediction. It encodes historical POI sequences and semantic information with a Transformer-based feature extractor, and further accounts for social influence and geographic constraints. 
Another line of research explores graph-based methods for next POI prediction, which construct graphs from mobility trajectories to capture user-POI interactions and POI transition patterns. These methods then apply GNNs to learn relational representations for predicting users' future POI visits \cite{STP-UDGAT,SGRec,Graph-flashback,STHGCN,rao2024next}. For instance, STP-UDGAT \cite{STP-UDGAT} captures POI-POI and user-user relationships with graph attention, combining personalized local preferences and global spatial-temporal-preference neighborhoods for next POI recommendation.
Graph-Flashback \cite{Graph-flashback} constructs a heterogeneous spatio-temporal knowledge graph to learn POI embeddings that capture transition patterns among POIs. STHGCN \cite{STHGCN} leverages a spatio-temporal hypergraph to model high-order dependencies and global collaborative relationships across mobility trajectories. It further integrates inter-user and intra-user collaborative signals with spatio-temporal contexts for next POI prediction. Despite their effectiveness in modeling sequential patterns and relational structures, these methods largely rely on observed mobility correlations and predefined contextual features, making it difficult to capture deeper behavioral intentions and semantic dependencies in check-in trajectories. This limits their generalization to diverse user behaviors and complex urban environments.

\textbf{LLM-based approaches.} Recent studies have explored LLMs for next POI prediction due to their strong reasoning and generation capabilities.
For example, LLM-Mob \cite{LLM-Mob} leverages the reasoning capabilities of LLMs through context-inclusive prompts that encode users' historical and recent mobility records, thereby capturing both long-term and short-term mobility dependencies. LLM-Move \cite{LLMmove} formulates next-POI prediction as a candidate ranking problem and introduces prompting strategies that incorporate geographic preferences, spatial distances, and sequential transition patterns. LLM4POI \cite{LLM4POI} adapts pretrained LLMs by prompt-based question answering, allowing the model to leverage contextual information and commonsense knowledge through fine-tuning.
Mobility-LLM \cite{Mobility-llm} extracts semantic information from check-in sequences to help LLMs better understand users' visiting intentions and travel preferences, and further fine-tunes the model on multiple mobility analysis tasks. POI-Enhancer \cite{Poi-enhancer} improves POI representations by leveraging LLM-derived semantic knowledge, while GNPR-SID \cite{GNPR-SID} constructs semantic POI identifiers from semantic and collaborative features for generative next POI recommendation. QT-Mob \cite{QT-Mob} adapts LLMs for mobility modeling by representing mobility records in textual space and encoding location semantics as discrete tokens, thereby capturing rich contextual information while remaining compatible with LLM architectures. SILO \cite{SILO} constructs a hybrid semantic space that integrates ID-based embeddings, context-based semantics, and auxiliary contextual information, enabling the joint modeling of sequential mobility patterns and rich contextual semantics. CoMaPOI \cite{Comapoi} introduces a collaborative multi-agent framework to address LLMs' limited understanding of numerical spatio-temporal data and alleviate irrelevant predictions caused by large candidate POI spaces.
However, existing LLM-based approaches often rely on prompt engineering or textualized trajectory representations. As a result, they may still struggle to jointly capture complex behavioral patterns, collaborative user relationships, and fine-grained travel intentions, thereby limiting their robustness and generalization.

\section{Preliminaries}
In this section, we elaborate the preliminary knowledge in the next POI prediction. 
Let $\mathcal{U}=\{ u_n \mid 1\leq n \leq N\}$ denote the set of $N$ users and $\mathcal{P}=\{p_m \mid 1\leq m\leq M\}$ denote the set of $M$ POIs. 
Each POI $p_m$ has a name, a category $c$, and a geographic location $(lat, lon)$, with $lat$ and $lon$ denoting the latitude and longitude respectively.

We are given $H$ historical trajectories of the user $u_n$, denoted as $\mathcal{X}_n= \{ X_n^1, X_n^2, \cdots, X_n^{H}\}$, where $X_n^i(1\le i\le H)$ denotes the $i$-th historical trajectory.
$X_n^i$ consists of several chronological check-in records, and we have $X_n^i=(x_{n,1}^i, x_{n,2}^i, \cdots)$, where $x_{n,j}^i$ represents the $j$-th check-in record in the $i$-th historical trajectory of $u_n$.
$x_{n,j}^i$ can be denoted as a tuple $(n, t_{i,j}, p)$, which respectively represents the user index, the visiting time, and the specific POI.



Given that the last check-in record in $X_n^i(1\le i\le H-1)$ is always earlier than the first one in $X_n^{i+1}$, we can reorganize the $H$ historical trajectories into one long trajectory for simplicity:
\begin{equation}
\begin{array}{c}
    X_n = (x_{n,1}^1, x_{n,2}^1, \cdots, x_{n,1}^2, x_{n,2}^2,, \cdots, x_{n,1}^H, x_{n,2}^H, \cdots) \\
    \Leftarrow \mathcal{X}_n = \{X_n^1, X_n^2, \cdots, X_n^H\}
\end{array}
\end{equation}

Given a \textit{query trajectory}~\footnote{We name a new sequence of check-in records that ends at the target time point as a query trajectory, which serves as the immediate context for the next POI prediction.} $X_n^{\rm q}=(x_{n,1}^{\rm q}, x_{n,2}^{\rm q}, \cdots)$ of the user $u_n$, we aim to predict the specific POI at the target time point $t_n^{\rm q}$, where $t_n^{\rm q}$ is later than the visiting time point of the last known record in $X_n^{\rm q}$.


\begin{figure*}[!t]
	\centering
    \includegraphics[width=\textwidth]{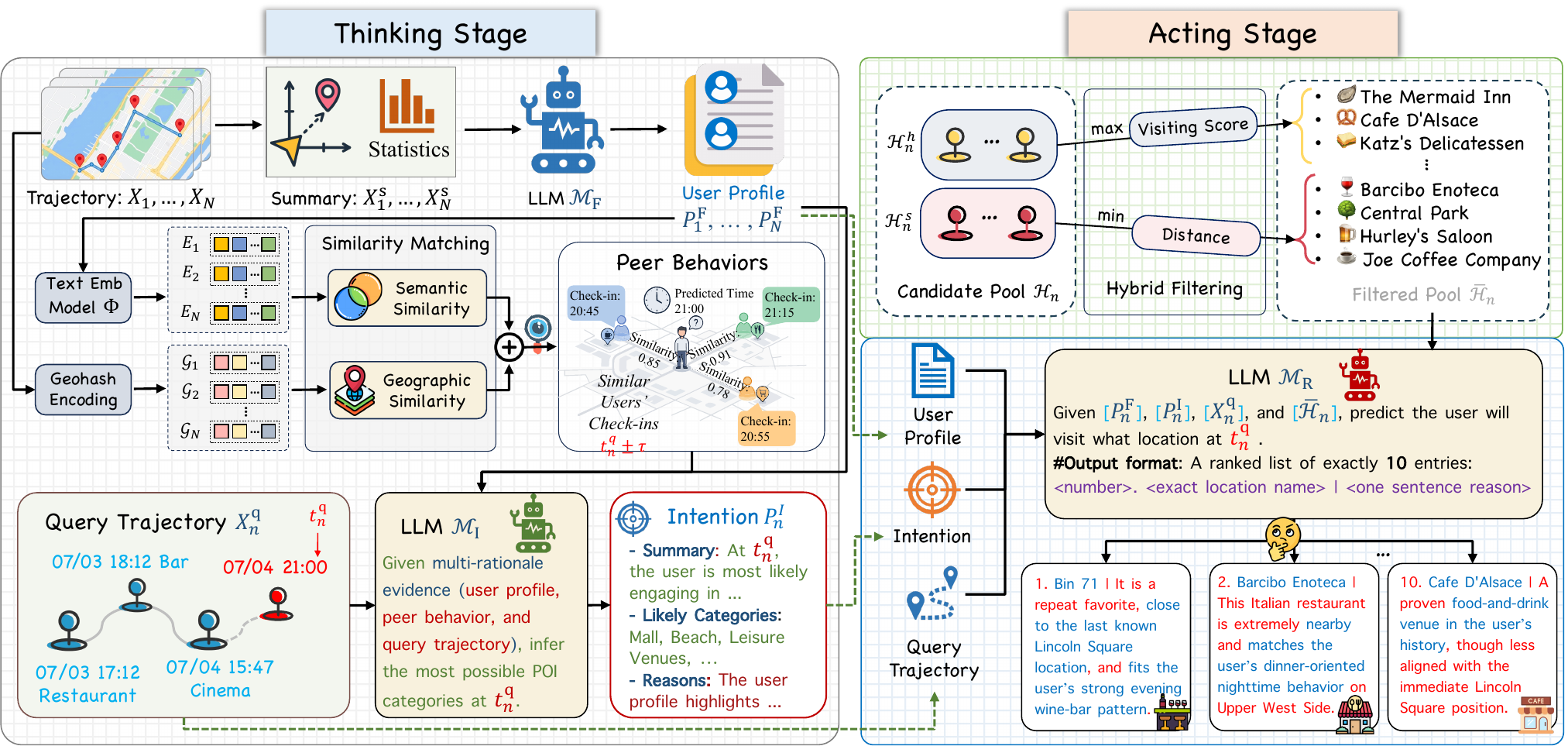}
	\caption{The workflow of \textbf{IntentPOI} includes Thinking Stage: Multi-Rationale Intention Inference and Acting Stage: Intention-Guided Reasoning for POI Prediction.}
	\label{method}
\end{figure*}

\section{Methodology}
The workflow of our proposed {IntentPOI} is presented in Fig.~\ref{method}.
The thinking stage infers latent intentions on multi-source rationale evidence, i.e., user profile, peer behaviors and temporal context. 
The acting stage firstly constructs a compact candidate pool, feed it to a reasoning LLM together with the inferred intention and supporting signals, where the intention serves as the reasoning scaffold for performing candidate recommendation.

\subsection{Thinking: Multi-Rationale Intention Inference}

Existing LLM-based methods collapse reasoning into a single implicit process and the LLM directly predicts the targeted POIs or re-rank candidates, without explicitly reasoning about the user's latent intention.
This heavily relies on the trajectory correlations and generates frequency-biased predictions.
Therefore, in this section, we aim to infer the intermediate intention to guide the downstream reasoning.
Specifically, we ground the intention on \textbf{multi-rationale evidence}, including \textit{user profile}, \textit{peer behaviors}, and the \textit{temporal context} from the query trajectory.


\textbf{User Profile.} 
The user profile can reflect the mobility patterns from the temporal and spatial perspective, providing a foundation for both intention inference and candidate recommendation.
We first extract a statistical summary from the historical trajectory $X_n$, capturing the frequency distributions of visiting hours, POIs, and POI categories.
Let $X_n^{\rm s}$ denote the statistical summary.
We then incorporate both $X_n^{\rm s}$ and $X_n$ into the prompt of a pretrained LLM $\mathcal{M}_{\rm F}$ to generate the profile of $u_n$:
\begin{equation}
    P_n^{\rm F} \gets \mathcal{M}_{\rm F} (X_n, X_n^{\rm s}),
\end{equation}
where $P_n^{\rm F}$ includes the inference of user's mobility pattern, e.g., \textit{a strong late-afternoon to evening rhythm} and \textit{an evening-oriented urban explorer}.
To avoid hallucination, $\mathcal{M}_{\rm F}$ is prompted to provide specific evidence to support the inference, based on the temporal and spatial analysis of the historical trajectory, as shown in Fig.~\ref{fig-showcase-profile-intention}(left).
The detailed prompt of $\mathcal{M}_{\rm F}$ is presented in Appendix.~\ref{sec_prompt}.

\textbf{Peer Behaviors.}
Given the insufficient contexts of sparse individual trajectories, peer behaviors can incorporate mobility evidence from similar users, thereby expanding the LLM's reasoning horizons from a single trajectory to multi-user evidence.
Users with similar mobility patterns tend to visit POIs of similar categories or even the same locations.
On the one hand, users active during similar temporal periods may have consistent latent intentions, such as dining or socializing, which may be invariant across geographic regions.
On the other hand, users with similar historical trajectories tend to share underlying mobility patterns, which increases the likelihood of visiting the same locations in the future.
Hence, we obtain both \textit{semantic} and \textit{geographic similarity} to evaluate the social connection among users.

We first obtain the semantic embedding of user's profile $P_n^{\rm F}$ by a text embedding model $\Phi$: 
\begin{equation}
    E_n = \Phi (P_n^{\rm F}).
\end{equation}
We then evaluate the semantic similarity of different users:
\begin{equation}
    s_{m,n}^{\rm E} = \mathbf{cos}(E_n, E_m) = \frac{E_n \cdot E_m}{\left\lVert E_n \right\rVert \left\lVert E_m \right\rVert},
    \label{eq-smt-sim}
\end{equation}
where $s_{m,n}^{\rm E}$ denotes the semantic similarity, i.e., how similar the mobility patterns of $u_n$ and $u_m$ is.

Each POI location $(lat, lon)$ is encoded into a geohash string at precision=5 using the Geohash algorithm~\cite{niemeyer2008geohash}, yielding cells of approximately $4.9 \times 4.9$ km. Each user's mobility footprint is then represented as a set of visited geohash cells:   
\begin{equation}
    \mathcal{G}_n = \{ \text{Geohash}(x_{n,i}(lat), x_{n,i}(lon)) \mid \forall x_{n,i} \in X_n\}
\end{equation}
The geographic similarity between two users is computed as the Jaccard index over their cell sets:
\begin{equation}
    s_{m,n}^{\rm G} = \frac{|\mathcal{G}_n \cap \mathcal{G}_m|}{|\mathcal{G}_n \cup \mathcal{G}_m|}. 
    \label{eq-geo-sim}
\end{equation}

We combine the semantic and geographic similarity with the balancing ratio $\alpha$ to obtain the final cross-user similarity:
\begin{equation}
    s_{m,n} = \alpha \cdot s_{m,n}^{\rm E} + (1-\alpha) \cdot s_{m,n}^{\rm G}.
    \label{eq-total-sim}
\end{equation}
Therefore we can obtain the similarity matrix $S=\{s_{m,n}\}_{1\leq m,n \leq N}$.
We then construct $\mathcal{U}_n$ by selecting the top-$k$ users with the highest similarity to $u_n$:
\begin{equation}
    \label{eq-social}
    \mathcal{U}_n = \{u_m \mid s_{m,n} \in \operatorname{arg\,top-k}_{1\leq m \leq N,m\neq n } s_{m,n} \}.
\end{equation}
We organize the visiting locations of these $k$ users in the time zone of $[t_n^{\rm q}-\tau , t_n^{\rm q}+\tau]$  as peer behaviors, denoted as $P_n^{\rm S}$.

\begin{figure*}[!t]
	\centering
    \includegraphics[width=\textwidth]{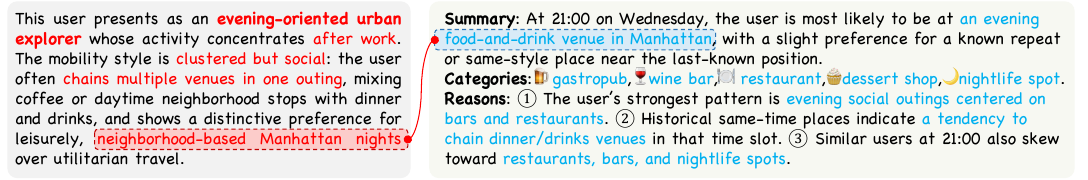}
	\caption{The user profile (left) concludes the mobility patterns with specific temporal and spatial habits. The inferred intention (right) includes the likely categories at $t_n^{\rm q}$ with multi-rationale evidence-grounded reasons from $P_n^{\rm F}$, $P_n^{\rm S}$, and $X_n^{\rm q}$.}
	\label{fig-showcase-profile-intention}
\end{figure*}
\textbf{Intention Inference.}
In existing LLM-based approaches, the reasoning LLM must simultaneously infer user intentions and rank candidate POIs, which is often limited to selecting the most frequent category rather than performing evidence-grounded reasoning.
We decouple these two tasks by pre-inferring a reasoned intention and using it as an explicit reasoning scaffold, thereby guiding the downstream LLM toward intention-aligned reasoning.

We prompt a LLM $\mathcal{M}_{\rm I}$ with rich contexts to generate reasonable intentions, which includes user profile, peer behaviors, and temporal contexts.
Given that users tend to revisit historical locations, the user profile $P_n^{\rm F}$ provides $\mathcal{M}_{\rm I}$ with both coarse-wise mobility patterns and fine-wise summary of the visiting frequency around $t_n^{\rm q}$.
The peer behaviors $P_n^{\rm S}$ enlarge the spatial horizons and provide $\mathcal{M}_{\rm I}$ with the locations of similar users.
The query trajectory $X_n^{\rm q}$ provides the temporal context and helps infer the travel purpose from the preceding check-in records.
We formulate the process of intention inference as:
\begin{equation}
    P_n^{\rm I} = \mathcal{M}_{\rm I} (X_n^{\rm q},P_n^{\rm F}, P_n^{\rm S}).
\end{equation}

$P_n^{\rm I}$ provides a structured description of the user's inferred traveling intention at the target time $t_n^{\rm q}$. 
As shown in Fig.~\ref{fig-showcase-profile-intention}(right), it captures the target POI category (e.g., \textit{Restaurant} or \textit{Wine Bar}) that best aligns with the user's temporal routine and profile patterns, together with a natural-language reasoning trace that grounds this intention in explicit evidence drawn from $P_n^{\rm F}$, $X_n^{\rm q}$, and $P_n^{\rm S}$. We provide the detailed prompt of $\mathcal{M}_{\rm I}$ in Appendix~\ref{sec_prompt}.

\subsection{Acting: Intention-Guided Reasoning for POI Prediction}

With the user intention built in the thinking stage, we perform intention-guided reasoning for candidate recommendation in this subsection.
Given that there are thousands of POIs in a city, directly prompting the LLM with all candidates is infeasible.
We therefore design a dual candidate selection strategy that first constructs a compact candidate pool and then applies hybrid filtering to retain the most promising candidates.
After filtering, the reasoning LLM evaluates candidates against the intention and all supporting signals, producing the final ranked recommendations.

\textbf{Candidate Pool Construction.} We construct the candidate pool from \textit{historical} and \textit{spatial} perspectives.
Firstly, given the fact that users tend to revisit familiar locations, we construct $\mathcal{H}_n^{\rm h}$ by including all historically-visited POIs.
In spatial perspective, we construct $\mathcal{H}_n^{\rm s}$, consisting of the $Z$ nearest POIs to the last known location in $X_n^{\rm q}$.
Therefore, we have the candidate pool $\mathcal{H}_n$ as
\begin{equation}
    \mathcal{H}_n = \mathcal{H}_n^{\rm h} \cup \mathcal{H}_n^{\rm s}.
\end{equation}
$\mathcal{H}_n$ provides a robust candidate pool that considers both visiting repetition and spatial proximity.

\textbf{Hybrid Filtering.}
To further narrow the candidate pool for efficient recommendation, we apply hybrid filtering from both historical and spatial perspectives to obtain the final candidate pool $\bar{\mathcal{H}}_n$ with $B$ candidates.
For each historical candidate $p \in \mathcal{H}_n^{\rm h}$, we calculate the visiting scores as: 
\begin{equation}
    L(p) = c_{\rm v} + c_{\rm d} + c_{\rm h},
\end{equation}
where $c_{\rm v}$ denotes the number of historical visits of $p$ by $u_n$; $c_{\rm d}$ denotes the number of historical visits on the same Day-of-Week with $t_n^{\rm q}$; $c_{\rm h}$ denotes the number of historical visits in the same hour buckets with $t_n^{\rm q}$.
We then select $\rho \times B$ candidates with highest visiting scores to construct $\bar{\mathcal{H}}_n^{\rm h}$.

In spatial perspective, we select $(1-\rho) \times B$ candidates with smallest distance to construct $\bar{\mathcal{H}}_n^{\rm s}$. 
Therefore, we can obtain the filtered candidate pool as $\bar{\mathcal{H}}_n = \bar{\mathcal{H}}_n^{\rm h} \cup \bar{\mathcal{H}}_n^{\rm s} $.

\textbf{Intention-Grounded Reasoning.}
In contrast to prior methods that feed raw trajectories or the intermediate user profiles into the LLM, 
we augment the LLM $\mathcal{M}_{\rm R}$ with the intention $P_n^{\rm I}$ to serve as the reasoning scaffold toward intention-aligned POI prediction.
We formulate the reasoning process as:
\begin{equation}
    \hat{\mathcal{Y}}_n = \mathcal{M}_{\rm R}\big(X_n^{\rm q}, P_n^{\rm F},\, P_n^{\rm I},\, \bar{\mathcal{H}}_n\big),
\end{equation}
where $\hat{\mathcal{Y}}_n = \{\hat{y}_1, \hat{y}_2, \ldots, \hat{y}_T\}$ denotes the ordered set of recommended POIs for user $u_n$ at the target time $t_n^{\rm q}$, given the query trajectory $X_n^{\rm q}$.
This grounded reasoning process ensures that each recommendation is supported by explicit and interpretable evidence rather than implicit trajectory patterns alone.

We present the overall process of {IntentPOI} in Algorithm~\ref{ag1} from the perspective of the operations in the historical trajectories (i.e, the training and validation sets in baselines) and query trajectories (i.e., the test sets in baselines).
We first build user profiles and evaluate users' similarity from historical trajectories, and then infer the intentions and the next POI results for the query trajectories. 
Therefore, we have a fair comparison with the baselines, without test data leakage.
More details are provided in Subsection~\ref{sec_settings}.

\begin{table*}[!th]
\centering
\tabcolsep=0.35cm

\caption{Comparison of different LLM-based methods for next POI prediction.}
\label{tab_comp}

\begin{tabularx}{\textwidth}{llll}
\toprule
{Dimensions} &
{Prompt-based Methods} &
{Token-based Methods} &
\textbf{IntentPOI} (Ours) \\
\midrule
Reasoning Paradigm
& One-step re-ranking
& One-step index mapping
& \textbf{Two-stage, \textcolor{blue}{Thinking $\rightarrow$ Acting}} \\
Intention Modeling
& Implicit
& Implicit
& \textbf{Explicit, \textcolor{blue}{intention as reasoning scaffold}} \\
Evidence Sources
& Historical-oriented
& Token semantics
& \textbf{Multi-source, \textcolor{blue}{including peer behaviors}} \\
Interpretability
& Limited
& Limited
& \textbf{High, \textcolor{blue}{evidence-grounded reasoning traces}}\\
\bottomrule
\end{tabularx}
\end{table*}

\begin{table}[!t]
\centering
\caption{Statistics of the datasets.}
\label{tab:dataset}
\tabcolsep=0.29cm
\begin{tabularx}{\columnwidth}{lcccc}
\toprule
Datasets & \#Users & \#POIs & \#Check-ins &\#Trajectories \\
\midrule
NYC & 1075 & 5,099 & 104,074 & 14,160 \\
TKY & 2,281 & 7,844 & 361,430 & 44,692 \\
CA  & 4,318 & 9,923 & 250,780 & 32,920 \\
\bottomrule
\end{tabularx}
\end{table}

\subsection{Comparative Analysis}

To further clarify the design rationale of {IntentPOI}, we provide an explicit comparison with two main LLM-based paradigms, i.e., prompt-based and token-based methods, as presented in Table~\ref{tab_comp}.

Prompt-based methods employ LLMs as one-step re-rankers~\cite{LLM-Mob, Comapoi, LLM4POI} and token-based methods similarly collapse reasoning into a single step, mapping token-encoded query trajectories directly to POI indices~\cite{QT-Mob, Poi-enhancer, GNPR-SID}.
\ding{202} Both paradigms ask the LLM to simultaneously infer what the user wants and which POI matches that need, without an explicit intermediate reasoning step.
In contrast, {IntentPOI} explicitly infers the intermediate intention, which serves as the reasoning scaffold for the LLM $\mathcal{M}_{\rm R}$ to yield intention-aligned recommendation results.
\ding{203} While both prompt-based and token-based methods derive evidence exclusively from a single user's trajectory patterns.
{IntentPOI} expands the evidence base to cross-user mobility patterns.
\ding{204} Moreover, in {IntentPOI}, each recommendation is supported by a full reasoning trace from $P_n^{\rm F}$, $P_n^{\rm I}$, and $X_n^{\rm q}$, making the decision process auditable.
In summary, our proposed IntentPOI shifts the LLM from a one-step trajectory-conditioned predictor to a two-stage intention-guided reasoner, where explicit intention inference bridges the gap between raw mobility signals and grounded location recommendations.

\begin{table*}[t]
\centering
\small
\tabcolsep=0.125cm
\caption{Performance comparison on NYC, TKY, and CA datasets. 
\textcolor{red}{\textbf{Bold}}: the best.
\textcolor{blue}{\underline{Underline}}: the second best.
``Improve'' denotes the relative improvement of \textbf{IntentPOI} over the best baseline.
}
\label{tab:main_results}

\begin{tabularx}{\textwidth}{lccccc ccccc ccccc}
\toprule


& \multicolumn{5}{c}{NYC}
& \multicolumn{5}{c}{TKY}
& \multicolumn{5}{c}{CA} \\

\cmidrule(lr){2-6}
\cmidrule(lr){7-11}
\cmidrule(lr){12-16}

\multirow{-2}{*}{Methods} 
& HR@5 & HR@10 & N@5 & N@10 & MRR
& HR@5 & HR@10 & N@5 & N@10 & MRR
& HR@5 & HR@10 & N@5 & N@10 & MRR \\

\midrule

\rowcolor{blue!3} DeepMove
& 0.3169 & 0.3651 & 0.2456 & 0.2614 & 0.2285
& 0.3855 & 0.4585 & 0.2965 & 0.3201 & 0.2767
& 0.2861 & 0.3377 & 0.2091 & 0.2257 & 0.1903 \\

\rowcolor{blue!3} GETNext
& 0.2973 & 0.3684 & 0.2111 & 0.2341 & 0.1921
& 0.1654 & 0.1934 & 0.1328 & 0.1419 & 0.1257
& 0.2153 & 0.2373 & 0.1637 & 0.1709 & 0.1497 \\

\rowcolor{blue!3} SASRec
& 0.2098 & 0.2365 & 0.1613 & 0.1701 & 0.1489
& 0.3338 & 0.3925 & 0.2494 & 0.2685 & 0.2294
& 0.1829 & 0.2146 & 0.1367 & 0.1470 & 0.1256 \\

\rowcolor{blue!3} BERT4Rec
& 0.2083 & 0.2461 & 0.1584 & 0.1708 & 0.1470
& 0.3272 & 0.4005 & 0.2436 & 0.2673 & 0.2256
& 0.1884 & 0.2256 & 0.1402 & 0.1520 & 0.1290 \\

\rowcolor{blue!3} FPMC
& 0.3010 & 0.3340 & 0.2296 & 0.2403 & 0.2102
& 0.4425 & 0.5261 & 0.3429 & 0.3700 & 0.3210
& 0.2503 & 0.3061 & 0.1935 & 0.2119 & 0.1823 \\

\rowcolor{blue!3} POI-GDE
& 0.2265 & 0.2624 & 0.1716 & 0.1832 & 0.1582
& 0.3388 & 0.4063 & 0.2584 & 0.2802 & 0.2406
& 0.1878 & 0.2283 & 0.1411 & 0.1542 & 0.1310 \\

\midrule

\rowcolor{green!3} AgentMove
& 0.4073 & 0.4893 & 0.3008 & 0.3272 & 0.2762
& 0.4240 & 0.5238 & 0.3075 & 0.3400 & 0.2823
& 0.3631 & 0.4305 & \textcolor{blue}{\underline{0.2878}} & \textcolor{blue}{\underline{0.3095}} & \textcolor{blue}{\underline{0.2716}} \\

\rowcolor{green!3} CoMaPOI
& 0.3643 & 0.4411 & 0.2670 & 0.2916 & 0.2448
& 0.3892 & 0.4881 & 0.2840 & 0.3160 & 0.2624
& \textcolor{blue}{\underline{0.3645}} & \textcolor{blue}{\underline{0.4580}} & 0.2653 & 0.2957 & 0.2451 \\

\rowcolor{green!3} LLM4POI
& 0.2620 & 0.3010 & 0.2023 & 0.2148 & 0.1876
& 0.3121 & 0.3602 & 0.2411 & 0.2567 & 0.2239
& 0.2442 & 0.2765 & 0.1905 & 0.2011 & 0.1770 \\

\rowcolor{green!3} MobilityLLM
& \textcolor{blue}{\underline{0.4075}} & \textcolor{blue}{\underline{0.4920}} & \textcolor{blue}{\underline{0.3112}} & \textcolor{blue}{\underline{0.3388}} & \textcolor{blue}{\underline{0.2978}}
& \textcolor{blue}{\underline{0.4519}} & \textcolor{blue}{\underline{0.5391}} & \textcolor{blue}{\underline{0.3431}} & \textcolor{blue}{\underline{0.3714}} & \textcolor{blue}{\underline{0.3276}}
& 0.3087 & 0.3799 & 0.2357 & 0.2586 & 0.2296 \\

\rowcolor{green!3} QT-Mob
& 0.3062 & 0.3358 & 0.2553 & 0.2650 & 0.2423
& 0.4005 & 0.4466 & 0.3273 & 0.3423 & 0.3092
& 0.2641 & 0.3095 & 0.2178 & 0.2328 & 0.2087 \\

\midrule

\rowcolor{red!3} \textbf{IntentPOI}
& \textcolor{red}{\textbf{0.4274}} & \textcolor{red}{\textbf{0.5063}} & \textcolor{red}{\textbf{0.3332}} & \textcolor{red}{\textbf{0.3586}} & \textcolor{red}{\textbf{0.3123}}
& \textcolor{red}{\textbf{0.5272}} & \textcolor{red}{\textbf{0.6265}} & \textcolor{red}{\textbf{0.3726}} & \textcolor{red}{\textbf{0.4045}} & \textcolor{red}{\textbf{0.3344}}
& \textcolor{red}{\textbf{0.4072}} & \textcolor{red}{\textbf{0.4851}} & \textcolor{red}{\textbf{0.3152}} & \textcolor{red}{\textbf{0.3372}} & \textcolor{red}{\textbf{0.2937}} \\

\midrule

\rowcolor{gray!5} Improve (\%)
& 4.88 & 2.91 & 7.07 & 5.84 & 4.87
& 16.66 & 16.21 & 8.60 & 8.91 & 2.08
& 11.71 & 5.92 & 9.52 & 8.95 & 8.14 \\

\bottomrule
\end{tabularx}
\end{table*}

\section{Experiments}
\subsection{Experimental Settings}
\label{sec_settings}

\textbf{Datasets.}
We conduct experiments on three widely-used real-world POI check-in datasets: NYC, TKY, and CA.
The NYC dataset includes check-in records collected in New York City from April 2012 to February 2013.
The TKY dataset covers Tokyo during the same period~\cite{yang2015nationtelescope}.
The CA dataset includes check-ins across California and Nevada, from February 2009 to October 2010~\cite{cho2011friendship}.
Following prior works~\cite{Comapoi,GETNext}, we filter out users and POIs with fewer than 10 check-ins.
The statistical details of the processed datasets are presented in Table~\ref{tab:dataset}.
For the baselines, these datasets are divided into training, validation, and test sets in chronological order with the ratio of 8:1:1. 
For fair comparison, in our proposed IntentPOI, we adopt the training and validation sets (i.e., the former 90\% trajectories) to build the user profiles and similarity matrix, and report the prediction performance on the test set.

\textbf{Baselines.}
We compare IntentPOI with a comprehensive collection of baselines, including \textbf{deep learning-based methods}: 
DeepMove~\cite{Deepmove}, GETNext~\cite{GETNext}, SASRec~\cite{SASRec}, BERT4Rec~\cite{BERT4Rec}, FPMC~\cite{rendle2010factorizing}, and POI-GDE~\cite{POIGDE}; 
and \textbf{LLM-based methods}: 
AgentMove~\cite{feng2025agentmove}, CoMaPOI~\cite{Comapoi}, LLM4POI~\cite{LLM4POI}, MobilityLLM~\cite{gong2024mobility}, and QT-Mob~\cite{QT-Mob}.


\textbf{Evaluation Metrics.}
Following prior works~\cite{Comapoi,GETNext,QT-Mob}, we adopt standard ranking-based metrics to evaluate the recommendation performance.
Hit Rate at $n$ (HR@$n$) measures the proportion of test cases where the ground-truth POI appears among the top-$n$ predictions.
Normalized Discounted Cumulative Gain at $n$ (N@$n$) further accounts for the ranking quality by assigning higher weights to correct predictions at top positions.
Mean Reciprocal Rank (MRR) evaluates the average reciprocal rank of the first correct answer.

\textbf{Implementation Details.}
We employ \texttt{GPT-5.4}\footnote{https://openai.com/index/introducing-gpt-5-4/} as $\mathcal{M}_{\rm F}$ and $\mathcal{M}_{\rm I}$ to generate user profiles and intentions respectively.
We employ \texttt{GPT-5.4-mini}\footnote{https://openai.com/index/introducing-gpt-5-4-mini-and-nano/} as $\mathcal{M}_{\rm R}$ to perform intention-guided reasoning.
We adopt \texttt{text-embedding-3-large}\footnote{https://openai.com/index/new-embedding-models-and-api-updates/} as the text embedding model $\Phi$.
The hyperparameters are set as follows: $k=5$ for peer selection, $\alpha=0.5$ for similarity fusion, $\tau=30$ min for temporal window, $Z=50$ for spatial candidate construction, $B=30$ for candidate pool size, $T=10$ for candidate recommendation, and $\rho=0.9$ for hybrid filtering ratio.
All experiments are conducted on a server with 4 NVIDIA A6000 GPUs.
The source code can be accessed online \footnote{https://github.com/yuppielqx/Next-POI}.

\subsection{Main Results}
We present the performance comparison on the three datasets in Table~\ref{tab:main_results}.
It is evident that {IntentPOI} consistently outperforms all baselines across all metrics and datasets, particularly on TKY and CA datasets, with HR@5 improved by $\mathbf{16.66}\%$ and $\mathbf{11.71}\%$ respectively.
The performance gains indicate that explicit intention reasoning provides a robust inductive bias that adapts to cities with different mobility patterns and POI densities.
LLM-based methods generally outperform deep learning-based approaches, indicating that the semantic understanding capabilities of LLMs are beneficial for modeling complex human mobility patterns.
Among the LLM-based baselines, MobilityLLM and AgentMove achieve competitive results.
However, both methods still operate as one-step predictors without explicit intention modeling, and their performance gap relative to {IntentPOI} increases on datasets with more diverse user behaviors, such as TKY and CA.

In conclusion, these results demonstrate the effectiveness of the proposed thinking-then-acting paradigm.
By explicitly inferring user intentions as an intermediate reasoning step, {IntentPOI} transforms next POI prediction from frequency-dominated pattern matching into intention-guided reasoning, leading to consistent and substantial improvements over state-of-the-art methods across diverse real-world scenarios.

\subsection{Model Analysis}

\textbf{Ablation in \textbf{IntentPOI}.}
We present five ablation variants of {IntentPOI} in Table~\ref{tab:ablation-config}, where each variant selectively removes one component from $\mathcal{M}_{\rm I}$ or $\mathcal{M}_{\rm R}$.
Moreover, we also ablate the process of candidate pool construction into three variants:
(i) adopting the random filtering strategy, (ii) including only spatial candidates, and (iii) including only historical candidates.

We report the ablation results in Table~\ref{tab:ablation-full}.
{IntentPOI} consistently outperforms all eight variants.
\ding{202} The impact of removing a signal depends critically on which LLM it feeds.
Ablating the user profile from $\mathcal{M}_{\rm I}$ results in remarkable performance drop, with the HR@1 decreasing by $49.8\%$ (from 0.211 to 0.106).
While removing the same profile signal from $\mathcal{M}_{\rm R}$ produces only a marginal decline, with the HR@1 decreasing by $3.4\%$.
This asymmetry directly validates that the user profile is the foundational evidence for the thinking stage, but the intention can compensate for its absence in the acting stage.
\ding{203} Removing the intention signal from $\mathcal{M}_{\rm R}$ results in $11.2\%$ HR@1 drop, indicating that the intention serves as the primary reasoning scaffold in the acting stage.
\ding{204} Removing peer behaviors from $\mathcal{M}_{\rm I}$ produces $9.5\%$ HR@1 degradation, indicating that cross-user evidence plays a supplementary role in intention inference.

\ding{205} Random filtering causes $20.4\%$ HR@1 drop, confirming that the hybrid filtering strategy effectively identifies relevant candidates for the final recommendation.
\ding{206} Removing historical candidates causes catastrophic collapse, with HR@1 decreasing from 0.211 to 0.010, confirming that re-visiting patterns dominate the POI prediction.
\ding{207} In contrast, removing spatial candidates leads to moderate degradation, with HR@1 decreasing by $2.4\%$, indicating that spatial proximity serves as a useful but secondary signal for the final recommendation.

\newcommand{\cmark}{\ding{52}}
\newcommand{\xmark}{\ding{56}}
\begin{table}[!t]
\centering
\tabcolsep=0.45cm
\caption{Ablation variants of \textbf{IntentPOI}.
}
\label{tab:ablation-config}
\begin{tabularx}{\columnwidth}{l c c c c}
\toprule
\multirow{2}{*}{Variants} & \multicolumn{2}{c}{$\mathcal{M}_{\rm I}$} & \multicolumn{2}{c}{$\mathcal{M}_{\rm R}$} \\
\cmidrule(lr){2-3} \cmidrule(lr){4-5}
& $P_n^{\rm F}$ & $P_n^{\rm S}$ & $P_n^{\rm F}$ & $P_n^{\rm I}$ \\
\midrule
\rowcolor{red!3}\textbf{IntentPOI} & \cmark & \cmark & \cmark& \cmark \\
\rowcolor{blue!3} A.1 (w/o $P_n^{\rm F}$ in $\mathcal{M}_{\rm I}$) & \textcolor{red}{\xmark} & \cmark & \cmark & \cmark \\
\rowcolor{blue!3} A.2  (w/o $P_n^{\rm S}$ in $\mathcal{M}_{\rm I}$)& \cmark & \textcolor{red}{\xmark} & \cmark & \cmark \\
\rowcolor{blue!3} A.3  (w/o $P_n^{\rm F}$ in $\mathcal{M}_{\rm R}$) & \cmark & \cmark & \textcolor{red}{\xmark} & \cmark \\
\rowcolor{blue!3} A.4  (w/o $P_n^{\rm I}$ in $\mathcal{M}_{\rm R}$) & \cmark & \cmark & \cmark & \textcolor{red}{\xmark} \\
\rowcolor{blue!3} A.5 (w/o All) & \textcolor{red}{\xmark} & \textcolor{red}{\xmark} & \textcolor{red}{\xmark} & \textcolor{red}{\xmark} \\
\bottomrule
\end{tabularx}
\end{table}

\begin{table}[!t]
\centering
\tabcolsep=0.08cm
\caption{Ablation study in \textbf{IntentPOI} on CA dataset.
\textcolor{best}{\textbf{Bold}}: the best. \textcolor{second}{\underline{Underline}}: the second best.}
\label{tab:ablation-full}
\begin{tabularx}{\columnwidth}{l c c c c c c c}
\toprule
Metrics
& HR@1
& HR@5
& HR@10
& N@1
& N@5
& N@10
& MRR \\
\midrule
\rowcolor{red!3}\textbf{IntentPOI}
& \textcolor{best}{\textbf{0.211}}
& \textcolor{best}{\textbf{0.407}}
& \textcolor{best}{\textbf{0.485}}
& \textcolor{best}{\textbf{0.211}}
& \textcolor{best}{\textbf{0.315}}
& \textcolor{best}{\textbf{0.337}}
& \textcolor{best}{\textbf{0.294}} \\
\rowcolor{blue!3}A.1 & 0.106 & 0.292 & 0.404 & 0.106 & 0.205 & 0.241 & 0.191 \\
\rowcolor{blue!3} A.2 & 0.191 & 0.397 & \textcolor{second}{\underline{0.475}}
& 0.191 & 0.302 & 0.331 & 0.282 \\
\rowcolor{blue!3} A.3 & 0.204 & 0.399 & 0.473 & 0.204 & 0.307 & 0.331 & 0.286 \\
\rowcolor{blue!3} A.4 & 0.186 & 0.361 & 0.455 & 0.186 & 0.277 & 0.306 & 0.261 \\
\rowcolor{blue!3} A.5 & 0.101 & 0.282 & 0.378 & 0.101 & 0.196 & 0.227 & 0.180 \\
\midrule
\rowcolor{green!3} random & 0.168 & 0.289 & 0.336 & 0.168 & 0.233 & 0.248 & 0.220 \\
\rowcolor{green!3} w/o $\mathcal{H}_n^{\rm s}$ & \textcolor{second}{\underline{0.206}} & \textcolor{second}{\underline{0.400}} & 0.471 & \textcolor{second}{\underline{0.206}} & \textcolor{second}{\underline{0.309}} & \textcolor{second}{\underline{0.333}} & \textcolor{second}{\underline{0.289}} \\
\rowcolor{green!3} w/o $\mathcal{H}_n^{\rm h}$
& 0.010 & 0.032 & 0.043 & 0.010 & 0.022 & 0.025 & 0.019 \\
\bottomrule
\end{tabularx}
\end{table}

\begin{figure}[!t]
	\centering
    \includegraphics[width=\columnwidth]{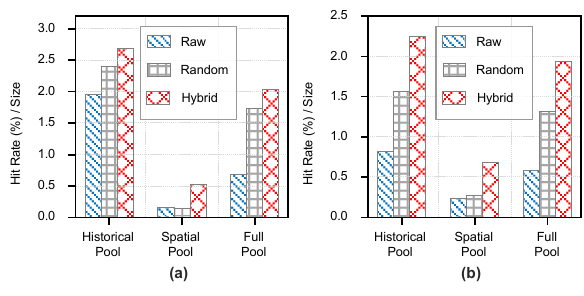}
	\caption{Coverage efficiency of different candidate filtering strategies on (a) NYC and (b) CA datasets.}
	\label{fig-filter-eff}
\end{figure}

We define a \textit{successful hit} as if the ground-truth POI is included in the candidate pool.
Therefore, we can calculate the average hit rate of the candidate pools across all query trajectories.
To further evaluate the efficiency of the hybrid filtering strategy, we introduce \emph{coverage efficiency} as the average hit rate in different filtering strategies divided by the pool size.
As shown in Fig.~\ref{fig-filter-eff}, our proposed hybrid filtering achieves the highest coverage efficiency in each candidate pool across the two datasets.
Specifically, it exceeds the raw full candidate pool (i.e., $\mathcal{H}_n$) by $3.0\times$ on NYC and $3.3\times$ on CA, and outperforms random filtering by $18\%$ and $47\%$ respectively on the full candidate pool.


Table~\ref{tab:ablation-sparse} further reports the results on sparse users (having less than 5 historical trajectories, i.e., $H \le 5$).
{IntentPOI} consistently outperforms the five variants.
Under sparse settings, the stage-dependent asymmetry persists but with a notable shift.
\ding{202} A.1 remains the most damaging ablation, indicating that the user profile is the foundational evidence for intention inference, and its importance is decreased when the historical trajectories offer sparse temporal contexts, with HR@1 decreasing by 42.72\% (from 0.206 to 0.118).
\ding{203} Comparing A.2 in the full-test and sparse settings, we have a key observation that the peer behaviors play a more critical role for sparse users, with the HR@1 decreasing by $14.6\%$ under sparse settings compared to $9.5\%$ in the full-test setting.
\ding{204} Moreover, the reasoning scaffold provided by intention becomes more critical for $\mathcal{M}_{\rm R}$ in sparse settings, compared with 11.2\% HR@1 decrease in Table~\ref{tab:ablation-full} and 18.93\% in Table~\ref{tab:ablation-sparse}.



\begin{table}[!t]
\centering
\tabcolsep=0.08cm
\caption{Ablation results on query trajectories of sparse users on CA dataset.
\textcolor{best}{\textbf{Bold}}: the best. \textcolor{second}{\underline{Underline}}: the second best.}
\label{tab:ablation-sparse}
\begin{tabularx}{\columnwidth}{l c c c c c c c}
\toprule
Metrics
& HR@1
& HR@5
& HR@10
& N@1
& N@5
& N@10
& MRR  \\
\midrule
\rowcolor{red!3}\textbf{IntentPOI}
& \textcolor{best}{\textbf{0.206}}
& \textcolor{best}{\textbf{0.398}}
& \textcolor{best}{\textbf{0.475}}
& \textcolor{best}{\textbf{0.206}}
& \textcolor{best}{\textbf{0.309}}
& \textcolor{best}{\textbf{0.334}}
& \textcolor{best}{\textbf{0.290}} \\
\rowcolor{blue!3} A.1 & 0.118 & 0.255 & 0.343 & 0.118 & 0.188 & 0.216 & 0.177 \\
\rowcolor{blue!3} A.2 & \textcolor{second}{\underline{0.176}} & \textcolor{second}{\underline{0.333}} & \textcolor{second}{\underline{0.412}} & \textcolor{second}{\underline{0.176}} & 0.261 & \textcolor{second}{\underline{0.286}} & \textcolor{second}{\underline{0.247}} \\
\rowcolor{blue!3} A.3 & 0.176 & 0.333 & 0.382 & 0.176 & \textcolor{second}{\underline{0.263}} & 0.277 & 0.244 \\
\rowcolor{blue!3} A.4 & 0.167 & 0.255 & 0.382 & 0.167 & 0.213 & 0.253 & 0.215 \\
\rowcolor{blue!3} A.5 & 0.098 & 0.235 & 0.294 & 0.098 & 0.166 & 0.185 & 0.152 \\
\bottomrule
\end{tabularx}
\end{table}

\begin{figure}[!t]
	\centering
    \includegraphics[width=\columnwidth]{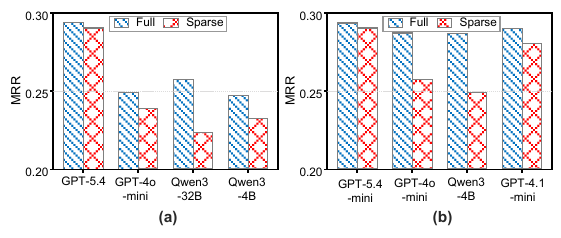}
	\caption{Full and sparse test performance in different implementations of (a) $\mathcal{M}_{\rm F}$, $\mathcal{M}_{\rm I}$, and (b) $\mathcal{M}_{\rm R}$.}
	\label{fig-llm-ab}
\end{figure}

\textbf{Ablation in Pretrained LLMs.}
Fig.~\ref{fig-llm-ab} reports MRR under varying selections of the generation LLMs ($\mathcal{M}_{\rm F}$ and $\mathcal{M}_{\rm I}$) in thinking stage and the reasoning LLM ($\mathcal{M}_{\rm R}$) in acting stage, in both full and sparse test settings.

{IntentPOI} is substantially more sensitive to the LLM choice for generation than for reasoning.
\ding{202} In Fig.~\ref{fig-llm-ab}(a), MRR spans a wide range across different implementations, which indicates that stronger signal generators produce more accurate user profiles and intentions, thus promoting the final prediction performance quality.
\ding{203} While in Fig.~\ref{fig-llm-ab}(b), the performance difference is relatively small. 
This asymmetry indicates that the thinking stage does the difficult reasoning to establish a high-quality intention, and the acting stage becomes a light evaluation task that even a moderate LLM can perform effectively.


\begin{figure}[!t]
	\centering
    \includegraphics[width=0.9\columnwidth]{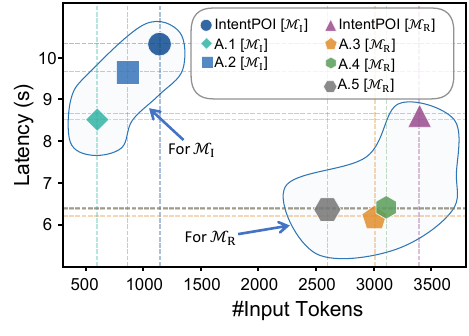}
	\caption{Inference cost of $\mathcal{M}_{\rm I}$ and $\mathcal{M}_{\rm R}$ across ablation variants in terms of input token count and latency.
    $\spadesuit [\heartsuit]$ denotes the inference cost of $\heartsuit$ in variant $\spadesuit$.
    }
	\label{fig-efficiency}
\end{figure}

\textbf{Efficiency Analysis.}
To assess the inference cost of each ablation variant in Table~\ref{tab:ablation-config}, we measure the input token count and latency of $\mathcal{M}_{\rm I}$ and $\mathcal{M}_{\rm R}$ on the CA dataset, averaged across all test queries.
Note that in Fig.~\ref{fig-efficiency}, we only report the measurements of LLMs which have ablation variants.

$\mathcal{M}_{\rm I}$ and $\mathcal{M}_{\rm R}$ exhibit a consistent cost asymmetry across all variants.
\ding{202} The input tokens of $\mathcal{M}_{\rm I}$ are relatively fewer and are prompted by the user profile, temporal contexts, and peer behaviors, resulting higher latency due to complicated intention generation.
\ding{203} $\mathcal{M}_{\rm R}$ consumes more input tokens for the candidate pool $\bar{\mathcal{H}}_n$ and all reasoning signals, but runs faster since it performs discriminative ranking over structured candidates rather than open-ended generation.
This indicates that the thinking stage performs a high-quality but concentrated reasoning step, while the acting stage performs a broader but mechanically lighter evaluation.
\ding{204} Removing individual signals can only yield marginal token savings for both LLMs, as the dominant token cost comes from the query trajectory and candidate pool, which remain constant across variants.
\ding{205} While A.5 eliminates the $\mathcal{M}_{\rm I}$ call and reduces $\mathcal{M}_{\rm R}$'s input tokens, this saving results in catastrophic accuracy collapse, as shown in Table~\ref{tab:ablation-full} and Table~\ref{tab:ablation-sparse}.
In conclusion, the efficiency analysis validates that the moderate inference cost of {IntentPOI} is necessary for the performance gains.

\begin{figure}[!t]
    \centering
    \includegraphics[width=\columnwidth]{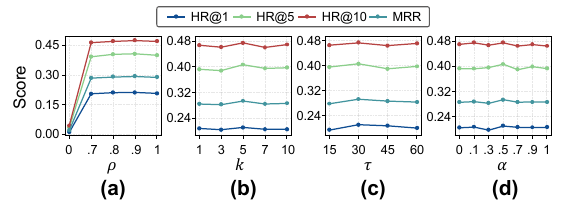}
    \caption{Performance under difference settings of 
    (a) historical candidate ratio $\rho$,
    (b) number of similar users $k$,
    (c) temporal window $\tau$ (minutes), and
    (d) semantic-geographic balancing ratio $\alpha$.}
    \label{fig-sensitivity}
\end{figure}

\textbf{Hyperparameter Investigation.}
We evaluate the performance variance of {IntentPOI} under different hyperparameter settings.
Fig.~\ref{fig-sensitivity} reports the numerical results of four metrics on CA dataset w.r.t four key hyperparameters, i.e., $\rho$, $k$, $\tau$, and $\alpha$.
Since the $Z$ spatial candidates are generated solely based on geographic distance, while only the top $(\rho \times B)$ candidates are retained after hybrid filtering, we do not investigate the impact of $Z$ on the final performance.

\begin{figure}[!t]
    \centering
    \includegraphics[width=\columnwidth]{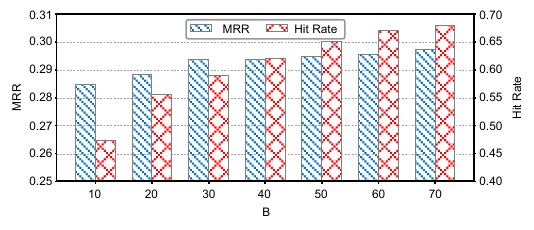}
    \caption{Effect of candidate pool size $B$ on MRR and average hit rate of the filtered candidate pool $\bar{\mathcal{H}}_n$ .}
    \label{fig-candi}
\end{figure}

\ding{202} In Fig.~\ref{fig-sensitivity}(a), the performance improves as $\rho$ increases, confirming that historical candidates are more informative than spatial candidates for next POI prediction, as they directly capture the user's re-visiting patterns.
However, the performance gain disappears after $\rho = 0.9$, indicating that spatial candidates compensate for the long-tail POIs that are not captured by historical candidates.

\ding{203} The candidate pool size $B$ affects the trade-off between accuracy and efficiency.
As shown in Fig.~\ref{fig-candi}, larger $B$ retains more candidates in the filtered pool $\bar{\mathcal{H}}_n$, increasing the hit rate of the ground truth, but also increases $\mathcal{M}_{\rm R}$'s input length and inference cost.
The default setting $B = 30$ achieves a good balance between the converged MRR and moderate token consumption.

\ding{204} We vary the number of similar users $k$ from 1 to 10, which affects the richness of the peer behaviors for intention inference.
As shown in Fig.~\ref{fig-sensitivity}(b), lower or higher values of $k$ may introduce noise or bias.
Too few peers may not provide representative samples of behavior patterns, while too many peers may include dissimilar users whose check-in patterns are less irrelevant to the target user.

\ding{205} Different values of $\tau$ affect the temporal relevance of the peer check-ins used for intention inference.
As shown in Fig.~\ref{fig-sensitivity}(c), the narrow temporal window ($\tau = 15$ min) may miss relevant check-ins from users whose mobility patterns are slightly offset from the target user's schedule, while too wide window ($\tau = 60$min) dilutes the temporal specificity of the peer evidence.

\ding{206} In Fig.~\ref{fig-sensitivity}(d), we vary the balancing ratio $\alpha$ from 0 (i.e., only geographic similarity) to 1 (i.e., only semantic similarity).
All metrics peak at $\alpha = 0.5$ and degrades with lower or higher values, confirming that the two similarity signals are complementary, with semantic similarity for mobility pattern alignment and geographic similarity for shared spatial footprints.

\ding{207} Across all four parameters, the default configuration is $\rho = 0.9$, $k = 20$, $\tau = 30$, and $\alpha = 0.5$.
All metrics vary smoothly within a reasonable range around each optimum rather than exhibiting sharp cliffs, which indicates that {IntentPOI} is not sensitive to hyperparameter selection, making the framework robust to dataset-specific tuning in practical applications.

\subsection{Case Study}

In this subsection, we showcase two samples from the CA dataset to illustrate the success of IntentPOI enhanced by the explicit intention and failure on the out-of-distribution trajectory.

\begin{figure}[!t]
	\centering
    \includegraphics[width=\columnwidth]{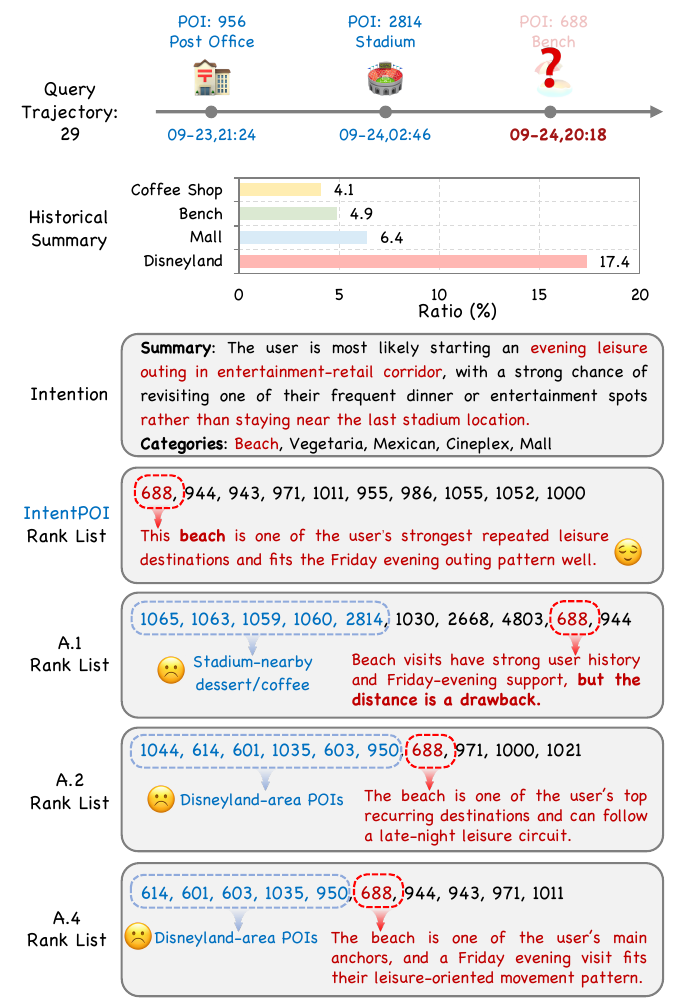}
	\caption{A successful case study on query trajectory~$\mathtt{29}$.
	The historical summary, inferred intention and reasoning details on four variants are provided.}
	\label{fig:case1}
\end{figure}

\textbf{Sucessful Case.}
We select the query trajectory~$\mathtt{29}$ from User~$\mathtt{9}$, which is a sparse case with 5 historical trips and the query trajectory has only 2 check-ins.
The user visits diverse categories and the top category (Disneyland Resort) accounts for only 17.4\% of visits, making it a challenging case where no single signal can dominate.
We compare the prediction results of three variants in Fig.~\ref{fig:case1}.

\begin{figure}[!t]
	\centering
    \includegraphics[width=\columnwidth]{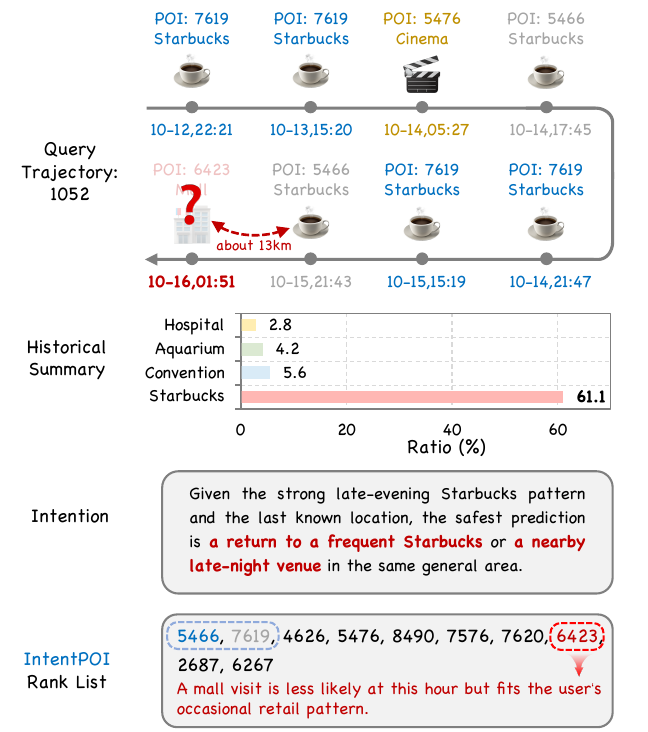}
	\caption{A failure case on query trajectory~$\mathtt{1052}$.
    The historical summary, inferred intention and reasoning details of \textbf{IntentPOI} are provided.}
	\label{fig:case2}
\end{figure}

The inferred intention in IntentPOI correctly identifies the user's likely activity as an \textit{evening leisure outing} instead of \textit{staying around stadium}, and constrains the candidate ranking to leisure-relevant categories, which is critical for correctly nominating the \textit{Beach} category.
This is based on the profile evidence of the user's leisure pattern in Orange County, the temporal context that Friday evening is a peak entertainment window, and the peer behavior evidence that similar users visit diverse non-Disneyland categories on Friday evenings.
Based on such, the {IntentPOI} correctly ranks the ground-truth Beach POI (POI~$\mathtt{688}$) at position~1.

The three ablation variants fail with distinct reasons.
\ding{202} A.1 (w/o $P_n^{\rm F}$ in $\mathcal{M}_{\rm I}$) fails to learn about the user's long-term mobility patterns, and the generated intention lacks spatial grounding.
Therefore, $\mathcal{M}_{\rm R}$ consequently defaults to pure spatial distance ranking, where all top-5 candidates are POIs near the Stadium, and the distant Beach POI falls to rank~9.
\ding{203} In A.2 (w/o $P_n^{\rm S}$ in $\mathcal{M}_{\rm I}$), the top-5 candidates all become the most-visited Disneyland Resort.
However, the user's own behavior is not Disneyland-dominated.
Without cross-user diversity evidence to counterbalance the raw frequency signal, the intention is inferred purely from the user's own history, and $\mathcal{M}_{\rm R}$ cannot distinguish between overall prevalence and time-specific relevance.
\ding{204} A.4 (w/o $P_n^{\rm I}$ in $\mathcal{M}_{\rm R}$) produces a superficially similar Disneyland collapse, but the underlying cause is distinct.
Without the intention scaffold, the temporal constraint ``09-24 20:18 $\rightarrow$ leisure'' is absent from $\mathcal{M}_{\rm R}$'s reasoning, and the LLM defaults to unconditional frequency ranking without category-level guidance.
\ding{205} Overall, the successful case indicates that multi-rationale signals in {IntentPOI} provide complementary rather than redundant reasoning priors, and that signal diversity is the key mechanism driving correct predictions.

\textbf{Failure Case.}
However, when the profile and peer behavior signals collectively fail to support the ground truth, the inferred intentions can hardly serve as an effective reasoning scaffold.
We present a representative failure case in Fig.~\ref{fig:case2} to illustrate this boundary.
We select query trajectory~$\mathtt{1052}$ (User~$\mathtt{363}$) from the CA dataset.
This user has 10 historical trajectories and 72 check-ins, with behavior overwhelmingly dominated by Starbucks (61.1\%).
The query trajectory places the user at two Starbucks locations in repeated alternation over six of the seven check-ins.
The ground-truth POI is a \textit{Mall} (ID: $\mathtt{6423}$) located 13~km north of the user's Starbucks cluster.
This instance exhibits three simultaneous deviations from the user's historical pattern.
(i)~Category: Mall accounts for only a single historical visit vs. 61.1\% Starbucks; 
(ii)~Temporal: the user has only 10 check-ins in the time scope of 00:00--02:00, scattered across 10 different POIs with no concentrated pattern; 
(iii)~Spatial: the ground-truth POI lies at the northern extreme of the user's activity range.

The user profile captures the user's Starbucks-dominated identity, 
but the Mall signal (only a single historical visit on a Monday midnight) is too weak.
The peer behavior retrieves similar users who at this hour predominantly visit Pubs, Bars, and Nightlife venues, instead of Malls.
Therefore, the intention can correctly identify the late-night context and predict likely categories as Starbucks, coffee shop, bar, and entertainment, but excludes Mall.
The final explanation of the Mall candidate is ``\textit{a mall visit is less likely at this hour but fits the user's occasional retail pattern}''.
The LLM simultaneously acknowledges the partial fit and the temporal implausibility, and with no signal providing positive support, this weak signal is overwhelmed by the Starbucks and nightlife evidence.

\section{Conclusion and Future Works}

Given the insight that users typically form an traveling intention before selecting a specific destination, we argue that the intention inference should be a critical intermediate reasoning step in next POI prediction task.
Therefore, we propose {IntentPOI}, a two-stage intention-guided reasoning framework that first infers user intentions from historical mobility patterns, peer behaviors, and query contexts, and then performs intention-guided POI recommendation. 
By explicitly incorporating intention as a reasoning scaffold, {IntentPOI} transforms next POI prediction into a more interpretable reasoning process. 
Extensive experiments on three real-world datasets demonstrate that {IntentPOI} consistently outperforms state-of-the-art baselines, while ablation studies verify the effectiveness of explicit intention reasoning.

\textbf{Limitations and Future Works.}
Despite its promising performance, {IntentPOI} still relies on LLM-generated intentions, whose quality may affect downstream recommendations. In addition, the current framework focuses on short-term intentions and does not explicitly model their evolution over time. Future work will explore intention-aware mobility modeling with more efficient models and investigate dynamic intention evolution for broader mobility prediction tasks.

\begin{acks}
We used ChatGPT to polish the sentences and improve the overall readability of the text.
Additionally, as a core component of the proposed IntentPOI, we utilized the \texttt{GPT-5.4} API as $\mathcal{M}_{\rm F}$ and $\mathcal{M}_{\rm I}$ to generate the user profiles and intentions, and the \texttt{GPT-5.4-mini} API as $\mathcal{M}_{\rm R}$ to infer the specific POIs.
\end{acks}

\bibliographystyle{ACM-Reference-Format}
\bibliography{software}

@String{Computing = "Computing" }

@String{Springer = "Springer-Verlag" }

@misc{niemeyer2008geohash,
    title  = {Geohash},
    author = {Niemeyer, Gustavo},
    year   = {2008},
    url    = {http://geohash.org}
  }

@article{10.1145/3510409,
author = {S\'{a}nchez, Pablo and Bellog\'{\i}n, Alejandro},
title = {Point-of-Interest Recommender Systems Based on Location-Based Social Networks: A Survey from an Experimental Perspective},
year = {2022},
issue_date = {January 2022},
publisher = {Association for Computing Machinery},
address = {New York, NY, USA},
volume = {54},
number = {11s},
issn = {0360-0300},
url = {https://doi.org/10.1145/3510409},
doi = {10.1145/3510409},
journal = {ACM Comput. Surv.},
month = sep,
articleno = {223},
numpages = {37}
}

@inproceedings{Deepmove,
  title={Deepmove: Predicting human mobility with attentional recurrent networks},
  author={Feng, Jie and Li, Yong and Zhang, Chao and Sun, Funing and Meng, Fanchao and Guo, Ang and Jin, Depeng},
  booktitle={The world wide web conference},
  pages={1459--1468},
  year={2018}
}

@inproceedings{Vanext,
  title={Predicting human mobility via variational attention},
  author={Gao, Qiang and Zhou, Fan and Trajcevski, Goce and Zhang, Kunpeng and Zhong, Ting and Zhang, Fengli},
  booktitle={The world wide web conference},
  pages={2750--2756},
  year={2019}
}

@inproceedings{Flashback,
  title={Location prediction over sparse user mobility traces using rnns},
  author={Yang, Dingqi and Fankhauser, Benjamin and Rosso, Paolo and Cudre-Mauroux, Philippe},
  booktitle={IJCAI},
  pages={2184--2190},
  year={2020}
}

@article{Mobtcast,
  title={MobTCast: Leveraging auxiliary trajectory forecasting for human mobility prediction},
  author={Xue, Hao and Salim, Flora and Ren, Yongli and Oliver, Nuria},
  journal={NeurIPS},
  volume={34},
  pages={30380--30391},
  year={2021}
}

@inproceedings{TSPN-RA,
  title={Towards effective next POI prediction: Spatial and semantic augmentation with remote sensing data},
  author={Jiang, Nan and Yuan, Haitao and Si, Jianing and Chen, Minxiao and Wang, Shangguang},
  booktitle={ICDE},
  pages={5061--5074},
  year={2024},
  organization={IEEE}
}

@article{LLM-Mob,
  title={Where would i go next? large language models as human mobility predictors},
  author={Wang, Xinglei and Fang, Meng and Zeng, Zichao and Cheng, Tao},
  journal={arXiv preprint arXiv:2308.15197},
  year={2023}
}

@inproceedings{LLMmove,
  title={Where to move next: Zero-shot generalization of llms for next poi recommendation},
  author={Feng, Shanshan and Lyu, Haoming and Li, Fan and Sun, Zhu and Chen, Caishun},
  booktitle={CAI},
  pages={1530--1535},
  year={2024},
  organization={IEEE}
}

@inproceedings{LLM4POI,
  title={Large language models for next point-of-interest recommendation},
  author={Li, Peibo and de Rijke, Maarten and Xue, Hao and Ao, Shuang and Song, Yang and Salim, Flora D},
  booktitle={SIGIR},
  pages={1463--1472},
  year={2024}
}

@inproceedings{Poi-enhancer,
  title={Poi-enhancer: An llm-based semantic enhancement framework for poi representation learning},
  author={Cheng, Jiawei and Wang, Jingyuan and Zhang, Yichuan and Ji, Jiahao and Zhu, Yuanshao and Zhang, Zhibo and Zhao, Xiangyu},
  booktitle={AAAI},
  volume={39},
  number={11},
  pages={11509--11517},
  year={2025}
}

@inproceedings{GNPR-SID,
  title={Generative Next POI Recommendation with Semantic ID},
  author={Wang, Dongsheng and Huang, Yuxi and Gao, Shen and Wang, Yifan and Huang, Chengrui and Shang, Shuo},
  booktitle={SIGKDD},
  pages={2904--2914},
  year={2025}
}

@inproceedings{Comapoi,
  title={Comapoi: A collaborative multi-agent framework for next poi prediction bridging the gap between trajectory and language},
  author={Zhong, Lin and Wang, Lingzhi and Yang, Xu and Liao, Qing},
  booktitle={SIGIR},
  pages={1768--1778},
  year={2025}
}

@inproceedings{QT-Mob,
  title={Enhancing large language models for mobility analytics with semantic location tokenization},
  author={Chen, Yile and Tao, Yicheng and Jiang, Yue and Liu, Shuai and Yu, Han and Cong, Gao},
  booktitle={SIGKDD},
  pages={262--273},
  year={2025}
}

@inproceedings{STP-UDGAT,
  title={STP-UDGAT: Spatial-temporal-preference user dimensional graph attention network for next POI recommendation},
  author={Lim, Nicholas and Hooi, Bryan and Ng, See-Kiong and Wang, Xueou and Goh, Yong Liang and Weng, Renrong and Varadarajan, Jagannadan},
  booktitle={Proceedings of the 29th ACM International conference on information \& knowledge management},
  pages={845--854},
  year={2020}
}

@inproceedings{Graph-flashback,
  title={Graph-flashback network for next location recommendation},
  author={Rao, Xuan and Chen, Lisi and Liu, Yong and Shang, Shuo and Yao, Bin and Han, Peng},
  booktitle={Proceedings of the 28th ACM SIGKDD conference on knowledge discovery and data mining},
  pages={1463--1471},
  year={2022}
}

@article{SGRec,
  title={Discovering collaborative signals for next POI recommendation with iterative Seq2Graph augmentation},
  author={Li, Yang and Chen, Tong and Luo, Yadan and Yin, Hongzhi and Huang, Zi},
  journal={arXiv preprint arXiv:2106.15814},
  year={2021}
}

@inproceedings{STHGCN,
  title={Spatio-temporal hypergraph learning for next POI recommendation},
  author={Yan, Xiaodong and Song, Tengwei and Jiao, Yifeng and He, Jianshan and Wang, Jiaotuan and Li, Ruopeng and Chu, Wei},
  booktitle={SIGIR},
  pages={403--412},
  year={2023}
}

@article{rao2024next,
  title={Next point-of-interest recommendation with adaptive graph contrastive learning},
  author={Rao, Xuan and Jiang, Renhe and Shang, Shuo and Chen, Lisi and Han, Peng and Yao, Bin and Kalnis, Panos},
  journal={TKDE},
  volume={37},
  number={3},
  pages={1366--1379},
  year={2024},
  publisher={IEEE}
}

@article{Mobility-llm,
  title={Mobility-llm: Learning visiting intentions and travel preference from human mobility data with large language models},
  author={Gong, Letian and Lin, Yan and Zhang, Xinyue and Lu, Yiwen and Han, Xuedi and Liu, Yichen and Guo, Shengnan and Lin, Youfang and Wan, Huaiyu},
  journal={Advances in Neural Information Processing Systems},
  volume={37},
  pages={36185--36217},
  year={2024}
}

@inproceedings{SILO,
  title={SILO: Semantic Integration for Location Prediction with Large Language Models},
  author={Sun, Tianao and Chen, Meng and Zhang, Bowen and Dai, Genan and Huang, Weiming and Zhao, Kai},
  booktitle={SIGKDD},
  pages={2756--2767},
  year={2025}
}

@article{yang2015nationtelescope,
  title={Modeling User Activity Preference by Leveraging User Spatial Temporal Characteristics in LBSNs},
  author={Yang, Dingqi and Zhang, Daqing and Zheng, Vincent W. and Yu, Zhiyong},
  journal={IEEE Transactions on Systems, Man, and Cybernetics: Systems},
  volume={45},
  number={1},
  pages={129--142},
  year={2014}
}

@inproceedings{cho2011friendship,
  title={Time-Aware Point-of-Interest Recommendation},
  author={Yuan, Quan and Cong, Gao and Ma, Zongyang and Sun, Aixin and Magnenat-Thalmann, Nadia},
  booktitle={SIGIR},
  pages={363--372},
  year={2013}
}

@inproceedings{SASRec,
  title={Self-Attentive Sequential Recommendation},
  author={Kang, Wang-Cheng and McAuley, Julian},
  booktitle={ICDM},
  pages={197--206},
  year={2018}
}

@inproceedings{BERT4Rec,
  title={BERT4Rec: Sequential Recommendation with Bidirectional Encoder Representations from Transformer},
  author={Sun, Fei and Liu, Jun and Wu, Jian and Pei, Changhua and Lin, Xiao and Ou, Wenwu and Jiang, Peng},
  booktitle={CIKM},
  pages={1441--1450},
  year={2019}
}

@inproceedings{GETNext,
  title={GETNext: Trajectory Flow Map Enhanced Transformer for Next POI Recommendation},
  author={Yang, Song and Liu, Jiamou and Zhao, Kaiqi},
  booktitle={SIGIR},
  pages={1144--1153},
  year={2022}
}

@article{POIGDE,
  title={Siamese Learning Based on Graph Differential Equation for Next-POI Recommendation},
  author={Yang, Yuxuan and Zhou, Siyuan and Weng, He and Wang, Dongjing and Zhang, Xin and Yu, Dongjin and Deng, Shuiguang},
  journal={Applied Soft Computing},
  volume={150},
  pages={111086},
  year={2024}
}

@article{luca2021survey,
  title={A survey on deep learning for human mobility},
  author={Luca, Massimiliano and Barlacchi, Gianni and Lepri, Bruno and Pappalardo, Luca},
  journal={ACM Computing Surveys (CSUR)},
  volume={55},
  number={1},
  pages={1--44},
  year={2021},
  publisher={ACM New York, NY}
}

@article{chen2025self,
  title={Self-supervised representation learning for geospatial objects: A survey},
  author={Chen, Yile and Huang, Weiming and Zhao, Kaiqi and Jiang, Yue and Cong, Gao},
  journal={Information Fusion},
  volume={123},
  pages={103265},
  year={2025},
  publisher={Elsevier}
}

@article{zhao2020go,
  title={Where to go next: A spatio-temporal gated network for next poi recommendation},
  author={Zhao, Pengpeng and Luo, Anjing and Liu, Yanchi and Xu, Jiajie and Li, Zhixu and Zhuang, Fuzhen and Sheng, Victor S and Zhou, Xiaofang},
  journal={IEEE Transactions on Knowledge and Data Engineering},
  volume={34},
  number={5},
  pages={2512--2524},
  year={2020},
  publisher={IEEE}
}

@inproceedings{lai2024disentangled,
  title={Disentangled contrastive hypergraph learning for next POI recommendation},
  author={Lai, Yantong and Su, Yijun and Wei, Lingwei and He, Tianqi and Wang, Haitao and Chen, Gaode and Zha, Daren and Liu, Qiang and Wang, Xingxing},
  booktitle={Proceedings of the 47th international ACM SIGIR conference on research and development in information retrieval},
  pages={1452--1462},
  year={2024}
}

@inproceedings{dang2023uniform,
  title={Uniform sequence better: Time interval aware data augmentation for sequential recommendation},
  author={Dang, Yizhou and Yang, Enneng and Guo, Guibing and Jiang, Linying and Wang, Xingwei and Xu, Xiaoxiao and Sun, Qinghui and Liu, Hong},
  booktitle={Proceedings of the AAAI conference on artificial intelligence},
  volume={37},
  number={4},
  pages={4225--4232},
  year={2023}
}

@inproceedings{yin2023next,
  title={Next POI recommendation with dynamic graph and explicit dependency},
  author={Yin, Feiyu and Liu, Yong and Shen, Zhiqi and Chen, Lisi and Shang, Shuo and Han, Peng},
  booktitle={Proceedings of the AAAI conference on artificial intelligence},
  volume={37},
  number={4},
  pages={4827--4834},
  year={2023}
}

@article{zeng2025global,
  title={Global and local hypergraph learning method with semantic enhancement for POI recommendation},
  author={Zeng, Jun and Tao, Hongjin and Tang, Haoran and Wen, Junhao and Gao, Min},
  journal={Information Processing \& Management},
  volume={62},
  number={1},
  pages={103868},
  year={2025},
  publisher={Elsevier}
}

@article{wu2025beyond,
  title={Beyond Regularity: Modeling Chaotic Mobility Patterns for Next Location Prediction},
  author={Wu, Yuqian and Peng, Yuhong and Yu, Jiapeng and Liu, Xiangyu and Yan, Zeting and Lin, Kang and Su, Weifeng and Qu, Bingqing and Lee, Raymond and Yang, Dingqi},
  journal={arXiv preprint arXiv:2509.11713},
  year={2025}
}

@inproceedings{wu2025mas4poi,
  title={Mas4poi: a multi-agents collaboration system for next poi recommendation},
  author={Wu, Yuqian and Peng, Yuhong and Yu, Jiapeng and Lee, Raymond},
  booktitle={Pacific-Asia Conference on Knowledge Discovery and Data Mining},
  pages={356--367},
  year={2025},
  organization={Springer}
}

@article{lv2026reasoning,
  title={Reasoning Over Space: Enabling Geographic Reasoning for LLM-Based Generative Next POI Recommendation},
  author={Lv, Dongyi and Ding, Qiuyu and Xu, Heng-Da and Sun, Zhaoxu and Wang, Zhi and Xiong, Feng and Xu, Mu},
  journal={arXiv preprint arXiv:2601.04562},
  year={2026}
}

@inproceedings{feng2018deepmove,
  title={Deepmove: Predicting human mobility with attentional recurrent networks},
  author={Feng, Jie and Li, Yong and Zhang, Chao and Sun, Funing and Meng, Fanchao and Guo, Ang and Jin, Depeng},
  booktitle={Proceedings of the 2018 world wide web conference},
  pages={1459--1468},
  year={2018}
}

@article{wu2020personalized,
  title={Personalized long-and short-term preference learning for next POI recommendation},
  author={Wu, Yuxia and Li, Ke and Zhao, Guoshuai and Qian, Xueming},
  journal={IEEE Transactions on Knowledge and Data Engineering},
  volume={34},
  number={4},
  pages={1944--1957},
  year={2020},
  publisher={IEEE}
}

@inproceedings{luo2021stan,
  title={Stan: Spatio-temporal attention network for next location recommendation},
  author={Luo, Yingtao and Liu, Qiang and Liu, Zhaocheng},
  booktitle={Proceedings of the web conference 2021},
  pages={2177--2185},
  year={2021}
}

@inproceedings{sun2024going,
  title={Going where, by whom, and at what time: Next location prediction considering user preference and temporal regularity},
  author={Sun, Tianao and Fu, Ke and Huang, Weiming and Zhao, Kai and Gong, Yongshun and Chen, Meng},
  booktitle={Proceedings of the 30th ACM SIGKDD Conference on Knowledge Discovery and Data Mining},
  pages={2784--2793},
  year={2024}
}

@inproceedings{yang2022getnext,
  title={Getnext: trajectory flow map enhanced transformer for next poi recommendation},
  author={Yang, Song and Liu, Jiamou and Zhao, Kaiqi},
  booktitle={Proceedings of the 45th International ACM SIGIR Conference on research and development in information retrieval},
  pages={1144--1153},
  year={2022}
}

@inproceedings{lim2020stp,
  title={STP-UDGAT: Spatial-temporal-preference user dimensional graph attention network for next POI recommendation},
  author={Lim, Nicholas and Hooi, Bryan and Ng, See-Kiong and Wang, Xueou and Goh, Yong Liang and Weng, Renrong and Varadarajan, Jagannadan},
  booktitle={Proceedings of the 29th ACM International conference on information \& knowledge management},
  pages={845--854},
  year={2020}
}

@article{wang2024embracing,
  title={Embracing LLMs for point-of-interest recommendations},
  author={Wang, Tianxing and Wang, Can},
  journal={IEEE Intelligent Systems},
  volume={39},
  number={1},
  pages={56--59},
  year={2024},
  publisher={IEEE}
}

@article{yang2024siamese,
  title={Siamese learning based on graph differential equation for Next-POI recommendation},
  author={Yang, Yuxuan and Zhou, Siyuan and Weng, He and Wang, Dongjing and Zhang, Xin and Yu, Dongjin and Deng, Shuiguang},
  journal={Applied Soft Computing},
  volume={150},
  pages={111086},
  year={2024},
  publisher={Elsevier}
}

@inproceedings{tan2024idgenrec,
  title={Idgenrec: Llm-recsys alignment with textual id learning},
  author={Tan, Juntao and Xu, Shuyuan and Hua, Wenyue and Ge, Yingqiang and Li, Zelong and Zhang, Yongfeng},
  booktitle={Proceedings of the 47th international ACM SIGIR conference on research and development in information retrieval},
  pages={355--364},
  year={2024}
}

@inproceedings{rendle2010factorizing,
  title={Factorizing personalized markov chains for next-basket recommendation},
  author={Rendle, Steffen and Freudenthaler, Christoph and Schmidt-Thieme, Lars},
  booktitle={Proceedings of the 19th international conference on World wide web},
  pages={811--820},
  year={2010}
}

@inproceedings{feng2025agentmove,
  title={Agentmove: A large language model based agentic framework for zero-shot next location prediction},
  author={Feng, Jie and Du, Yuwei and Zhao, Jie and Li, Yong},
  booktitle={Proceedings of the 2025 Conference of the Nations of the Americas Chapter of the Association for Computational Linguistics: Human Language Technologies (Volume 1: Long Papers)},
  pages={1322--1338},
  year={2025}
}

@article{gong2024mobility,
  title={Mobility-llm: Learning visiting intentions and travel preference from human mobility data with large language models},
  author={Gong, Letian and Lin, Yan and Zhang, Xinyue and Lu, Yiwen and Han, Xuedi and Liu, Yichen and Guo, Shengnan and Lin, Youfang and Wan, Huaiyu},
  journal={Advances in Neural Information Processing Systems},
  volume={37},
  pages={36185--36217},
  year={2024}
}

\appendix

\section{Overall Process of \textbf{IntentPOI}}
\label{sec_alg}
The overall process of {IntentPOI} is presented in Algorithm~\ref{ag1}, which consists of yielding user profiles and similarity matrix from historical trajectories (highlighted in \textcolor{blue}{blue}) and inference phase for the query trajectory.

For each user $u_n$, we generate the user profile $P_n^{\rm F}$ (Lines~2--3) and the pairwise user similarity matrix $S = \{s_{m,n}\}$ by fusing semantic and geographic similarities (Lines~4--6). The results are stored and reused across all queries.
For each query trajectory, we construct the peer behavior $P_n^{\rm S}$ by selecting the top-$k$ similar users and retrieve their check-in records within the temporal window $t^q_n \pm \tau$ (Lines~7--8). 
The intention $P_n^{\rm I}$ is then generated by $\mathcal{M}_{\rm I}$ (Line~9). 
For candidate selection, we construct a pool $\mathcal{H}_n$ from historical and spatial perspectives and apply hybrid filtering (Lines~11--14). Finally, the reasoning LLM $\mathcal{M}_{\rm R}$ performs intention-grounded reasoning to produce the top-$T$ recommendations $\hat{\mathcal{Y}}_n$ (Line~15).

\section{Detailed Prompts}
\label{sec_prompt}
We provide the detailed prompts of $\mathcal{M}_{\rm F}$ and $\mathcal{M}_{\rm I}$ in the right two figures.

\begin{algorithm}[!b]
    \caption{\textbf{IntentPOI}: Training and Inference}
    \label{ag1}
    \LinesNumbered
    \KwIn{
        $\{X_n\}_{n=1}^N$: historical trajectories of $N$ users; \\
        Query: $(u_n, t^q_n, X^q_n)$; \\
        Pretrained LLMs: $\mathcal{M}_{\rm F}, \mathcal{M}_{\rm I}, \mathcal{M}_{\rm R}$; \\
        Hyperparameters: $k, \alpha, \tau, Z, B, \rho, T$.
    }
    \KwOut{Recommended POIs $\hat{\mathcal{Y}}_n$}

    \For{$n \in [1, N]$}{
        {\color{blue} $X_n^{\rm s} \gets$ statistical summary of $X_n$\;}
        {\color{blue} $P_n^{\rm F} \gets \mathcal{M}_{\rm F}(X_n, X_n^{\rm s})$ \tcp*{\textbf{user profile}}}
        {\color{blue} $\mathcal{G}_n = \{ \text{Geohash}(x_{n,i}(lat), x_{n,i}(lon)) \mid \forall x_{n,i} \in X_n\}$ } \;
        {\color{blue} $E_n \gets \Phi(P_n^{\rm F})$ ;}
    }
    {\color{blue} $S \gets \{s_{m,n}\}_{N \times N}$, with $s_{m,n}$ computed via Eq.~~(\ref{eq-smt-sim}),~(\ref{eq-geo-sim}),~(\ref{eq-total-sim})}

    $\mathcal{U}_n = \{u_m \mid s_{m,n} \in \operatorname{arg\,top-k}_{1\leq m \leq N,m\neq n } s_{m,n} \}$ \;
    $P_n^{\rm S} \gets$ check-ins of $u_m \in \mathcal{U}_n$ in $t^q_n \pm \tau$ \tcp*{peer behaviors}
    $P_n^{\rm I} \gets \mathcal{M}_{\rm I} (X^q_n,\textcolor{blue}{P_n^{\rm F}}, P_n^{\rm S})$ \tcp*{intention}
    $\mathcal{H}_n^{\rm h} \gets \{p \mid (n, t, p) \in$ {\color{blue} $X_n$}$\}$ \;
    $\mathcal{H}_n^{\rm s} \gets Z$ nearest POIs to last location in $X^q_n$ \;
    $\mathcal{H}_n \gets \mathcal{H}_n^{\rm h} \cup \mathcal{H}_n^{\rm s}$ \tcp*{candidate pool}
    $\bar{\mathcal{H}}_n \gets$ filter $\mathcal{H}_n$  \tcp*{hybrid filtering}

    $\hat{\mathcal{Y}}_n \gets \mathcal{M}_{\rm R}(X_n^q,${\color{blue} $P_n^{\rm F}$}$, P_n^{\rm I}, \bar{\mathcal{H}}_n)$ \tcp*{intention-grounded reasoning}
    \Return $\hat{\mathcal{Y}}_n$
\end{algorithm}

\begin{figure}[!b]
    \label{prompt_f}
    \begin{tcolorbox}[colback=blue!5!white, 
                  colframe=blue!75!black, 
                  colbacktitle=blue!25!white, 
                  coltitle=black,         
                  fonttitle=\bfseries,   
                  title=1. Prompt in $\mathcal{M}_{\rm F}$ for user profiles,        
                  boxrule=1pt,           
                  arc=6pt,               
                  left=0.5mm,                  
                  right=0.5mm,                 
                  top=0.5mm,
                  bottom=0.5mm
                  ]    

    \textbf{\underline{System Prompt}}:

    You are a mobility analyst. Based on the following data about a Foursquare user in \textit{New York City}, write a comprehensive user profile (2-3 paragraphs, approximately 200 words).

    \textbf{\underline{User Prompt}}:

    \textbf{\#\# Statistical Summary}

    The user with ID 1 has taken 6 trips in total.
    \begin{itemize}[leftmargin=*]
        \item The top hours and frequencies for this user are: 18:00-18:59 for 4 times; 16:00-16:59 for 3 times...
        \item The top locations and frequencies are: `P.J. Clarke's' for 2 times; `La Colombe Torrefaction' for 2 times...
        \item The top categories and frequencies are: `bar' for 2 times; `gastropub' for 2 times...
    \end{itemize}

    \textbf{\#\# Visit History}
    \begin{itemize}[leftmargin=*]
        \item 2012-04-08 16:02 | Hi-Life Bar \& Grill | bar
        \item 2012-04-09 12:20 | Bubby's | popular American restaurant
        \item $\cdots$
    \end{itemize}

    

  Write in third person, present tense. Be specific — reference actual venue names and neighborhoods where possible. Do NOT include bullet points or headers; write flowing paragraphs.
\end{tcolorbox}
\end{figure}

\begin{figure}[!b]
    \label{prompt_i}
    \begin{tcolorbox}[colback=blue!5!white, 
                  colframe=blue!75!black, 
                  colbacktitle=blue!25!white, 
                  coltitle=black,         
                  fonttitle=\bfseries,   
                  title=2. Prompt in $\mathcal{M}_{\rm I}$ for intention inference,        
                  boxrule=1pt,           
                  arc=6pt,               
                  left=0.5mm,                  
                  right=0.5mm,                 
                  top=0.5mm,
                  bottom=0.5mm
                  ]    

    \textbf{\underline{System Prompt}}:

    You predict where a \textit{New York City} user is likely to be at a known future time. Return only JSON.

    \textbf{\underline{User Prompt}}:

    \textbf{\#\# User Profile}

    This user presents as an evening-oriented urban explorer whose activity concentrates after work and into the night ...
    
    \textbf{\#\# Temporal Contexts}
    \begin{itemize}[leftmargin=*]
        \item 17:12 on 2012-07-03: BLT Fish Shack (American restaurant)
        \item 18:12 on 2012-07-03: Rye House (Whisky Bar)
        \item 15:47 on 2012-07-04: AMC Loews Lincoln Square 13 (cinema) ← LAST KNOWN LOCATION
    \end{itemize}
    The next visit to predict happens on 2012-07-04 (Wednesday) at 21:00 during the night. Use this as a prior, but do not assume the destination.

    \textbf{\#\# Similar Users Did Around The Target Time ($\pm$30 min)}
    \begin{itemize}[leftmargin=*]
        \item - Similar user (similarity 0.86): around 21:00, was at Medi Winebar, then visited Thalia.
        \item - Similar user (similarity 0.84): around 21:00, was at Ditch Plains, then visited Westside Market.
        \item $\cdots$
    \end{itemize}

    Based on this trajectory and the known target time, infer where the user is most likely to be at that moment.
  Return JSON with exactly these keys: ``summary'', ``likely\_categories'', and ``rationale''.
\end{tcolorbox}
\end{figure}

\end{document}